\documentclass{bmvc2k}

\title{Adversarial Training for Multi-Channel Sign Language Production}

\addauthor{Ben Saunders}{b.saunders@surrey.ac.uk}{1}
\addauthor{Necati Cihan Camg\"{o}z}{n.camgoz@surrey.ac.uk}{1}
\addauthor{Richard Bowden}{r.bowden@surrey.ac.uk}{1}

\addinstitution{
 Centre for Vision, Speech \\ and Signal Processing\\
 University of Surrey\\
 Guildford, UK
}

\runninghead{Saunders \etal}{Adversarial Training for Multi-Channel SLP}

\def\etal{\emph{et al}\bmvaOneDot}

\usepackage[nolist]{acronym}
\begin{acronym}[PHOENIX14\textbf{T} ]

\acrodefplural{slrt}[SLRTs]{Sign Language Recognition Transformers}
\acrodefplural{sltt}[SLTTs]{Sign Language Translation Transformers}

\acro{slrt}[SLRT]{Sign Language Recognition Transformer}
\acro{sltt}[SLTT]{Sign Language Translation Transformer}

\acrodefplural{rnn}[RNNs]{Recurrent Neural Networks}
\acrodefplural{cnn}[CNNs]{Convolutional Neural Networks}
\acrodefplural{hmm}[HMMs]{Hidden Markov Models}
\acrodefplural{gru}[GRUs]{Gated Recurrent Units}
\acrodefplural{crf}[CRFs]{Conditional Random Fields}
\acrodefplural{gan}[GANs]{Generative Adversarial Networks}
\acrodefplural{gpu}[GPUs]{Graphic Processing Units}
\acro{bsl}[BSL]{British Sign Language}
\acro{bleu}[BLEU]{Bilingual Evaluation Understudy}
\acro{blstm}[BLSTM]{Bidirectional Long Short-Term Memory}
\acro{cnn}[CNN]{Convolutional Neural Network}
\acro{crf}[CRF]{Conditional Random Field}
\acro{cslr}[CSLR]{Continuous Sign Language Recognition}
\acro{ctc}[CTC]{Connectionist Temporal Classification}
\acro{dl}[DL]{Deep Learning}
\acro{dgs}[DGS]{German Sign Language - Deutsche Gebärdensprache}
\acro{dsgs}[DSGS]{Swiss German Sign Language - Deutschschweizer Geb\"ardensprache}
\acro{dtw}[DTW]{Dynamic Time Warping}
\acro{fc}[FC]{Fully Connected}
\acro{ff}[FF]{Feed Forward}
\acro{gan}[GAN]{Generative Adversarial Network}
\acro{gpu}[GPU]{Graphics Processing Unit}
\acro{gru}[GRU]{Gated Recurrent Unit}
\acro{gtpt}[G2PT]{Gloss to Pose Transformer}
\acro{hmm}[HMM]{Hidden Markov Model}
\acro{isl}[ISL]{Irish Sign Language}
\acro{lstm}[LSTM]{Long Short-Term Memory}
\acro{mha}[MHA]{Multi-Headed Attention}
\acro{mtc}[MTC]{Monocular Total Capture}
\acro{mse}[MSE]{Mean Squared Error}
\acro{nmt}[NMT]{Neural Machine Translation}
\acro{nlp}[NLP]{Natural Language Processing}
\acro{ph12}[PHOENIX12]{RWTH-PHOENIX-Weather-2012}
\acro{ph14}[PHOENIX14]{RWTH-PHOENIX-Weather-2014}
\acro{ph14t}[PHOENIX14\textbf{T}]{RWTH-PHOENIX-Weather-2014\textbf{T}}
\acro{pof}[POF]{Part Orientation Field}
\acro{paf}[PAF]{Part Affinity Field}
\acro{pttt}[P2TT]{Pose to Text Transformer}
\acro{relu}[RELU]{Rectified Linear Units}
\acro{rnn}[RNN]{Recurrent Neural Network}
\acro{rouge}[ROUGE]{Recall-Oriented Understudy for Gisting Evaluation}
\acro{sgd}[SGD]{Stochastic Gradient Descent}
\acro{sla}[SLA]{Sign Language Assessment}
\acro{slr}[SLR]{Sign Language Recognition}
\acro{slt}[SLT]{Sign Language Translation}
\acro{slp}[SLP]{Sign Language Production}
\acro{smt}[SMT]{Statistical Machine Translation}

\acro{ttgt}[T2GT]{Text to Gloss Transformer}
\acro{ttpt}[T2PT]{Text to Pose Transformer}
\acro{ttp}[T2P]{Text to Pose}
\acro{ttgtp}[T2G2P]{Text to Gloss to Pose}
\acro{wer}[WER]{Word Error Rate}

\end{acronym}
\usepackage{todonotes}
\usepackage{amsmath,amssymb}
\usepackage{color}
\usepackage{soul}
\usepackage{booktabs}
\usepackage{tablefootnote}
\usepackage{microtype}
\def\B{\fontseries{b}\selectfont}

\begin{document}

\maketitle




\begin{abstract}

Sign Languages are rich multi-channel languages, requiring articulation of both manual (hands) and non-manual (face and body) features in a precise, intricate manner. \ac{slp}, the automatic translation from spoken to sign languages, must embody this full sign morphology to be truly understandable by the Deaf community. Previous work has mainly focused on manual feature production, with an under-articulated output caused by regression to the mean.

In this paper, we propose an \textit{Adversarial Multi-Channel} approach to \ac{slp}. We frame sign production as a minimax game between a transformer-based \textit{Generator} and a conditional \textit{Discriminator}. Our adversarial discriminator evaluates the realism of sign production conditioned on the source text, pushing the generator towards a realistic and articulate output. Additionally, we fully encapsulate sign articulators with the inclusion of non-manual features, producing facial features and mouthing patterns.

We evaluate on the challenging \ac{ph14t} dataset, and report state-of-the art \ac{slp} back-translation performance for manual production. We set new benchmarks for the production of multi-channel sign to underpin future research into realistic \ac{slp}.

\end{abstract}

\section{Introduction} \label{sec:intro}

Sign languages, the principal communication of the Deaf community, are rich multi-channel languages. Communication is expressed through manual articulations of hand shape and motion, in combination with diverse non-manual features including mouth gestures, facial expressions and body pose \cite{sutton1999linguistics}. The combination of manual and non-manual features is subtle and complicated, requiring a detailed articulation to fully express the desired meaning. \acf{slp}, the translation from spoken language input to sign language output, is therefore required to encompass the full sign morphology in order to generate an accurate and understandable production.

Although sign languages are inherently multi-channel languages, deep learning based \ac{slp} approaches have, to date, focused solely on the manual features of sign \cite{saunders2020progressive,stoll2020text2sign,zelinka2020neural}, producing only the hand and body articulators. Ignoring non-manual features discards the contextual and grammatical information that is required to fully understand the meaning of the produced sign \cite{valli2000linguistics}. Mouthing, in particular, is vital to the comprehension of most sign languages, differentiating signs that may otherwise be homophones. Previous \ac{slp} models have also been trained using a regression loss \cite{saunders2020progressive,zelinka2020neural}, which results in an under-articulated production due to the problem of regression to the mean. Specifically, an average sign pose is generated, with non-expressive hand shape and body motion.

\begin{figure}[t!]
    \centering
    \includegraphics[width=0.99 \linewidth]{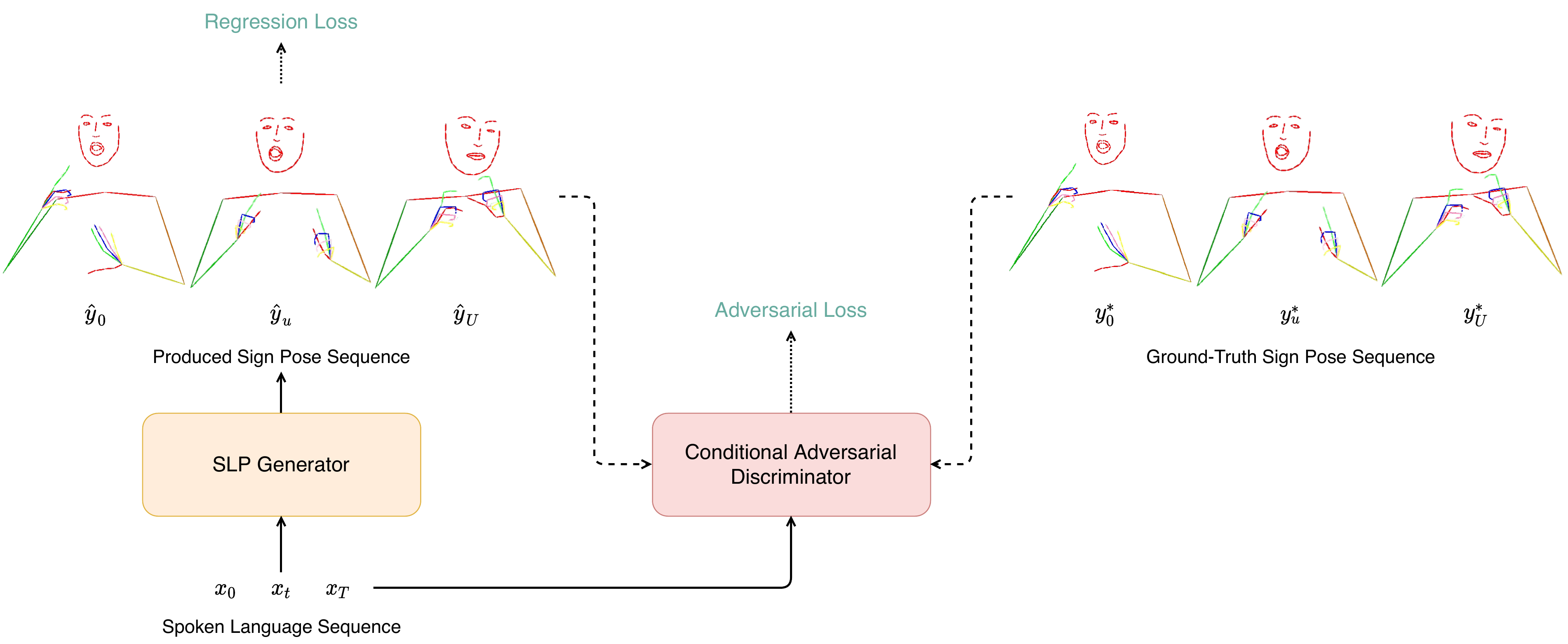}
    \caption{Adversarial Multi-Channel \ac{slp} overview, with a Conditional Adversarial Discriminator measuring the realism of Sign Pose Sequences produced by an SLP Generator.}
    \label{fig:adversarial_overview}
\end{figure}

In this paper, we propose adversarial training for multi-channel \ac{slp}, implementing a discriminator model conditioned on the source spoken language sentence, and expanding production to non-manual features. We frame \ac{slp} as a minimax game between a progressive transformer \textbf{Generator} that produces a sequence of sign poses from input text, and a conditional \textbf{Discriminator} that evaluates and promotes the realism of sign production. Building on the increase in discriminative production, we expand \ac{slp} to include \textbf{Non-Manual Features}, producing the head motion and mouthing patterns alongside the hands and body for a more expressive output. An overview of our approach is shown in Figure~\ref{fig:adversarial_overview}.

We evaluate on the \acf{ph14t} dataset using a back translation evaluation, achieving state-of-the-art results for the production of manual features and setting new benchmarks for non-manual and multi-channel production. We provide qualitative examples, demonstrating the impact of adversarial training in increasing the articulation of sign production.
The contributions of this paper can be summarised as:
\begin{itemize}
    \itemsep0em
    \item The first application of conditional adversarial training to \ac{slp}, to produce expressive and articulate sign pose sequences
    \item The first \ac{slp} model to fully encapsulate sign articulators through the production of non-manual features
    \item State-of-the-art \ac{slp} results on the \ac{ph14t} dataset, with baselines for multi-channel sign production
\end{itemize}
The rest of this paper is organised as follows: We outline the previous work in \ac{slp} and adversarial training in Section~\ref{sec:related_work}, and the background on machine translation and transformer models in Section~\ref{sec:background}. We present our Adversarial Multi-Channel approach for \ac{slp} in Section~\ref{sec:methodology}, with quantitative and qualitative evaluation provided in Section~\ref{sec:quant_experiments}. Finally, we conclude the paper in Section~\ref{sec:conclusion} by discussing our findings and future work. 

\section{Related Work} \label{sec:related_work}

\paragraph{Sign Language Recognition \& Translation}

The goal of vision-based sign language research is to develop systems capable of recognition, translation and production of sign languages \cite{bragg2019sign}. Although studied for the last three decades \cite{starner1997real,tamura1988recognition}, previous work has mainly focused on \ac{slr} \cite{camgoz2017subunets,koller2016deepsign,koller2019weakly}. These early works relied on manual features to understand sign, but as further linguistic aspects of sign were understood \cite{wilbur2000phonological,pfau2010nonmanuals}, focus shifted to more than just the hands. Subsequent tackling of the modalities of face \cite{vogler2008facial,koller2015continuous}, head pose \cite{luzardo2013head} and mouthings \cite{antonakos2012unsupervised,koller2015deep} have aided recognition performance.

Recently, Camgoz \etal introduced the first end-to-end \ac{slt} approach \cite{camgoz2018neural}, learning a translation from sign videos to spoken language rather than just recognising the sequence of signs. \ac{slt} is more demanding than \ac{slr} due to sign language possessing different linguistic rules and grammatical syntax from spoken language \cite{stokoe1980sign}. \ac{nmt} networks are predominantly used in \ac{slt} \cite{camgoz2018neural,ko2019neural,yin2020sign}, translating directly to spoken language or via gloss\footnote{Glosses are a written representation of sign language, and defined as minimal lexical items.} intermediary. Transformer based models are the state-of-the-art in \ac{slt}, jointly learning the recognition and translation tasks \cite{camgoz2020sign,camgoz2020multi}.

\paragraph{Sign Language Production}

Previous work into \ac{slp} has focused on avatar-based \cite{bangham2000virtual,kipp2011sign,lu2011data,zwitserlood2004synthetic} or \ac{smt} \cite{kayahan2019hybrid,kouremenos2018statistical} methods, requiring expensive motion capture or post-processing, with output limited to pre-recorded phrases. Non-manual features have been included in avatar production, such as mouthings \cite{wolfeexploring} and head positions \cite{cox2002tessa}, but are often viewed as ``stiff and emotionless'' with an ``absence of mouth patterns'' \cite{kipp2011assessing}.

More recently, there have been approaches to automatic \ac{slp} via deep learning \cite{xiao2020skeleton,zelinka2020neural}. However, these works focus on the production of isolated signs of a set length and order without realistic transitions, resulting in robotic and non-realistic animations that are poorly received by the Deaf \cite{bragg2019sign}. Stoll \etal \cite{stoll2018sign,stoll2020text2sign} use \acp{gan} to generate a sign language video of a human signer, as opposed to a skeleton pose. Even though the output video is visually pleasing, the approach still relies on the concatenation of isolated signs, which disregards the grammatical syntax of sign.

The closest work to this paper is that of Saunders \etal \cite{saunders2020progressive}, who use a progressive transformer architecture to produce continuous 3D sign pose sequences, utilising a counter decoding to predict sequence length and drive generation. However, the use of regression-based training, even with multiple data augmentation techniques, suffers from the known problem of regression to the mean, resulting in an under-expressed sign production. 

All previous deep learning based \ac{slp} works produce only manual features, ignoring the important non-manuals. The expansion to non-manual features is challenging due to the requirement of temporal coherence with manual features and the intricacies of facial movement. We expand production to non-manual features via the use of adversarial training to synchronise manual features and produce natural, expressive sign. 

\paragraph{Adversarial Training}

Since being introduced by Goodfellow \etal \cite{goodfellow2014generative}, \acp{gan} have been used extensively to generate images of increasing realism, pairing a generator and discriminator model in an adversarial training setup. \acp{gan} have produced impressive results when applied to image generation \cite{radford2015unsupervised,zhu2017unpaired,isola2017image} and, more recently, video generation tasks \cite{vondrick2016generating,tulyakov2018mocogan}. Conditional \acp{gan} \cite{mirza2014conditional} extend \acp{gan} to a dependent setting, enabling generation conditioned on specific external data inputs.

There has been recent progress in using \acp{gan} for natural language tasks \cite{lin2017adversarial,yu2017seqgan,zhang2016generating}. Specific to \ac{nmt}, Wu \etal designed Adversarial-NMT \cite{wu2017adversarial}, complimenting the original \ac{nmt} model with a \ac{cnn} based adversary, and Yang \etal \cite{yang2017improving} proposed a \ac{gan} setup with translation conditioned on the input sequence.

Specific to human pose generation, adversarial discriminators have been used for the production of realistic pose videos \cite{cai2018deep,chan2019everybody,ren2019music}. Ginosar \etal show that the task of generating skeleton motion suffers from regression to the mean, and adding an adversarial discriminator can improve the realism of gesture production \cite{ginosar2019learning}. Lee \etal utilise a conditioned discriminator to produce smooth and diverse human dancing motion from music \cite{lee2019dancing}. 

\section{Background} \label{sec:background}

In this section, we provide a brief background on \ac{nmt} sequence-to-sequence models, focusing on the recent transformer networks and their application to \ac{slp}. The goal of machine translation is to learn the conditional probability $P(Y|X)$ of generating a target sequence $Y = (y_1, ..., y_U)$ of $U$ tokens, given a source sequence $X = (x_1, ..., x_T)$ with $T$ tokens.

\acp{rnn} were first introduced for sequence-to-sequence tasks, mapping between sequences of different lengths using an iterative hidden state computation \cite{kalchbrenner2013recurrent}. The encoder-decoder architecture was later developed, encoding the source sentence into a ``context'' vector used to decode the target sequence \cite{cho2014learning,sutskever2014sequence}. However, this context introduced an information bottleneck and long term dependency issues. Attention mechanisms overcame this by expanding the context to a soft-search over the entire source sequence, conditioning each target prediction with a learnt weighting of the encoded tokens \cite{bahdanau2015neural,luong2015effective}.

Building on attention mechanisms, Vaswani \etal introduced the transformer network, a feed-forward model that replaces recurrent modules with self-attention and positional encoding \cite{vaswani2017attention}. Within each encoder and decoder stack, \ac{mha} layers perform multiple projections of self-attention, learning complementary representations of each sequence. The decoder utilises a further \ac{mha} sub-layer to combine these representations, learning the mapping between source and target sequences in an auto-regressive manner.
\paragraph{Progressive Transformer Model}
Sign languages are inherently continuous, encompassing fluid motions of hand shape, body pose and facial expressions. As \ac{slp} represents sign with continuous joint positions \cite{stoll2018sign,zelinka2020neural}, classic symbolic \ac{nmt} architectures, such as transformers, cannot be applied directly without modification. To tackle this, Saunders \etal proposed a progressive transformer architecture \cite{saunders2020progressive}, an alternative formulation of transformer decoding for continuous sequences. The model employs a counter decoding mechanism that drives generation and enables a prediction of the sequence end, alleviating the need for the classic end of sequence token found in symbolic \ac{nmt}. Multiple \ac{mha} sub-layers are applied over both the source, $x_{1:T}$, and target, $y_{1:U}$, sequences separately, with a final \ac{mha} layer used to learn the translation mapping between them. This can be formalised as:

\begin{equation}
\label{eq:progressive_transformer}
    [\hat{y}_{u+1},\hat{c}_{u+1}] = \textrm{ProgressiveTransformer} (y_{u}  \mid y_{1:u-1} , x_{1:T})
\end{equation}
where $\hat{y}_{u+1}$ and $\hat{c}_{u+1}$ are the produced joint positions and counter value respectively, given the source sentence, $x_{1:T}$, and previously predicted target poses, $y_{1:u-1}$. The model can be trained end-to-end using a regression loss of \ac{mse} between the ground truth, $y_{i}^{*}$, and produced, $\hat{y}_{i}$, sign pose sequences:
\begin{equation} \label{eq:loss_mse}
    \mathcal{L}_{Reg} = \frac{1}{U} \sum_{i=1}^{U} ( y_{i}^{*} - \hat{y}_{i} ) ^{2}
\end{equation}

In this paper, we build upon the progressive transformer architecture, employing a conditional adversarial discriminator that supplements the regression loss with an adversarial loss. This mitigates the effect of regression to the mean and prediction drift found in the original architecture. To further improve sign comprehension, we also include production of the non-manual sign features of facial expressions and mouthings.

\section{Adversarial Training for Multi-Channel \ac{slp}} \label{sec:methodology}

In this section, we introduce our \textbf{Adversarial Training} scheme for \textbf{Multi-Channel \ac{slp}}, learning to distinguish between real and fake sign pose sequences to ensure the production of realistic and expressive multi-modal sign language. Our objective is to learn a conditional probability $P(Y|X)$ of generating a target sign pose sequence $Y = (y_1, ..., y_U)$ of $U$ time steps, given a source spoken language sentence $X = (x_1, ..., x_T)$ with $T$ words. 

Realistic sign consists of subtle and precise movements of both manuals and non-manuals. However, \ac{slp} models often suffer from regression to the mean resulting in under-articulated output, producing average hand shapes due to the high variability of joint positions. To address the under-articulation of sign production, we propose an adversarial training mechanism for \ac{slp}. We utilise the previously described progressive transformer architecture (Section~\ref{sec:background}) as a \textbf{Generator}, $G$, to produce sign pose sequences from input text. To ensure realistic and expressive sign production, we introduce a conditional adversarial \textbf{Discriminator}, $D$, which learns to differentiate real and generated sign pose conditioned on the input spoken language. These models are co-trained in an adversarial manner, with mutually improved performance. The adversarial training scheme for \ac{slp} can thus be formalised as a minimax game, with $G$ aiming to minimise the following equation, whilst $D$ maximises it:
\begin{equation} \label{eq:loss_gan}
    \min_{G} \max_{D} \mathcal{L}_{GAN}(G,D) = \mathbb{E} [\log D(Y^{*} \mid X)] + \mathbb{E} [\log (1-D(G(X) \mid X))] 
\end{equation}
where $Y^{*} = y_{1:U}^{*}$ is the ground truth sign pose sequence, $G(X)$ equates to the produced sign pose sequence, $\hat{Y} = \hat{y}_{1:U}$, and $X$ is the source spoken language.

In addition to the adversarial training, we incorporate \textbf{Non-Manual Feature} production to create a more realistic signer output. Non-manual features are essential in the understanding of sign language, providing grammatical syntax, context and emphasis \cite{pfau2010nonmanuals}. In this paper, we model the facial landmarks of the signer, expanding sign pose sequences, $Y$, to include head nods, mouthings and eyebrow motion. The facial landmarks of a signer can be represented as coordinates, similar to the manuals, enabling a direct regression. 

\subsection{Generator}

Our \textbf{Generator}, $G$, learns to produce sign pose sequences given a source spoken language sequence, integrating the progressive transformer into a \ac{gan} framework. Contrary to the standard \ac{gan} implementation, we require sequence generation to be conditioned on a specific source input. Therefore, we remove the traditional noise input \cite{goodfellow2014generative}, and generate a sign pose sequence conditioned on the source sequence, taking inspiration from conditional \acp{gan} \cite{mirza2014conditional}.

We propose training $G$ using a combination of loss functions, namely regression loss, $\mathcal{L}_{Reg}$ (Equation~\ref{eq:loss_mse}), and adversarial loss, $\mathcal{L}^{G}_{GAN}$ (Equation~\ref{eq:loss_gan}), with the total loss function as:
\begin{equation} \label{eq:loss_total}
    \mathcal{L}^{G} = \lambda_{Reg} \mathcal{L}_{Reg}(G) + \lambda_{GAN} \mathcal{L}^{G}_{GAN}(G,D) 
\end{equation}
where $\mathcal{L}^{G}_{GAN}$ is the latter component of Equation~\ref{eq:loss_gan} and $\lambda_{Reg}$, $\lambda_{GAN}$ determines the importance of each loss function during training. The regression loss provides specific details about how to produce the given input, whereas the adversarial loss ensures a realistic signer motion. These losses work in tandem to create both an accurate and expressive sign production.

\subsection{Discriminator}
\begin{figure}[t!]
    \centering
    \includegraphics[width=1.0 \linewidth]{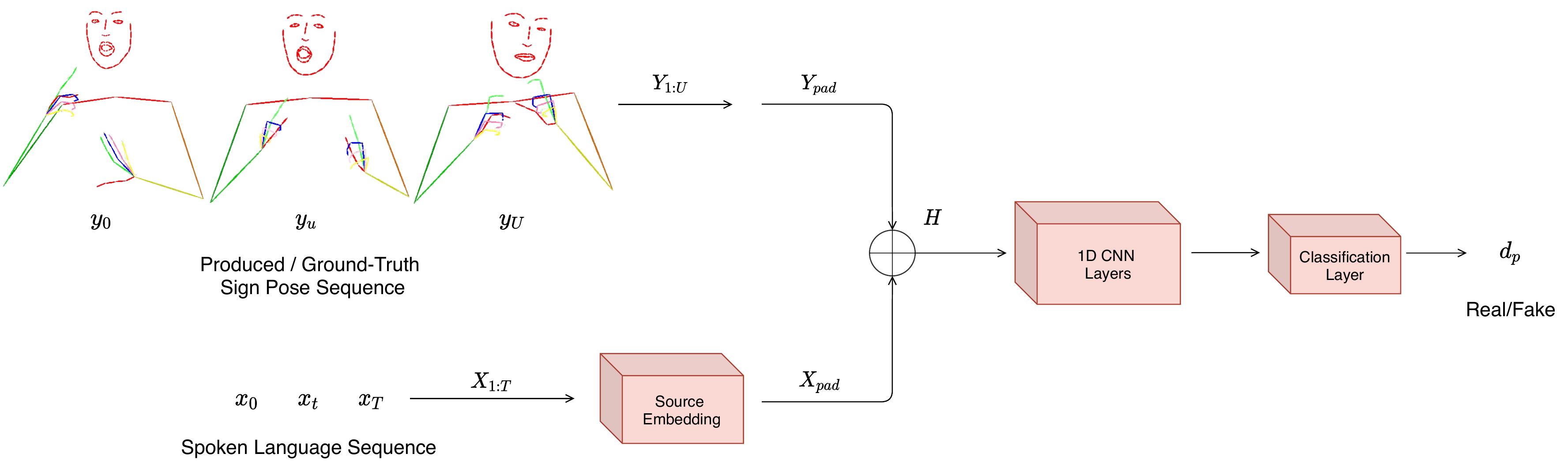}
    \caption{Architecture details of our Conditional Adversarial Discriminator. Sign pose, $Y_{1:U}$, is concatenated with source spoken language, $X_{1:T}$, and projected to a single scalar, $d_{p}$.}
    \label{fig:discriminator}
\end{figure}%
We present a conditional adversarial \textbf{Discriminator}, $D$, used to differentiate generated sign sequences, $\hat{Y}$, and ground-truth sign sequences, $Y^{*}$, conditioned on the source spoken language sequence, $X$. The aim of $D$ is to measure the realism of sign production, prompting $G$ towards an expressive and articulate output. In parallel, conditioning on the source sequence allows $D$ to concurrently measure the translation accuracy of source-target sequence pair, $(X,Y)$. Figure~\ref{fig:discriminator} shows an overview of the discriminator architecture.

For each pair of source-target sequences, $(X,Y)$, of either generated or real sign pose, the aim of the discriminator is to produce a single scalar, $d_{p} \in (0,1)$, representing the probability that the sign pose sequence originates from the data, $Y^{*}$:
\begin{equation} \label{eq:discriminator_scalar}
    d_{p} = P(Y = Y^{*} \mid X,Y) \in (0,1)  
\end{equation}
Due to the variable frame lengths of the sign sequences, we apply padding to transform them to a fixed length, $U_{max}$, the maximum frame length of target sequences found in the data:
\begin{equation} \label{eq:target_pad}
    Y_{pad} = [Y_{1:U} , \varnothing_{U:U_{max}}]
\end{equation}
where $Y_{pad}$ is the sign pose sequence padded with zero vectors, $\varnothing$, enabling convolutions upon the now fixed size tensor. In order to condition the discriminator on the source spoken language, we first embed the source tokens via a linear embedding layer. Again dealing with variable sequence lengths, these embeddings are also padded to a fixed length $T_{max}$, the maximum source sequence length:
\begin{equation} \label{eq:source_pad}
    X_{pad} = [W^{X} \cdot X_{1:T} + b^{X}, \varnothing_{T:T_{max}}]
\end{equation}
where $W^{X}$ and $b^{X}$ are the weight and bias of the source embedding respectively and $\varnothing$ is zero padding. As shown in the centre of Figure~\ref{fig:discriminator}, the source representation is then concatenated with the padded sign pose sequence, to create the conditioned features, $H$:
\begin{equation} \label{eq:discriminator_input}
    H = [Y_{pad}, X_{pad}]
\end{equation}
To determine the realism of the sign pose sequence, the discriminator extracts meaningful representations through multiple 1D \ac{cnn} layers. These convolutional filters are passed over the sign pose at the sequence level, analysing the local context to determine the temporal continuity of the signing motion. This is more effective than a frame level discriminator at determining realism, as a mean hand shape is a valid pose for a single frame, but not consistently over a large temporal window. Leaky ReLU activation \cite{maas2013rectifier} is applied after each layer, promoting healthy gradients during training. A final feed-forward linear layer and sigmoid activation projects the combined features down to the single scalar, $d_{p}$, representing the probability that the sign pose sequence is real.

We train the discriminator by maximising the likelihood of producing $d_{p} = 1$ for real sign sequences and $d_{p} = 0$ for generated sequences. This objective can be formalised as maximising Equation~\ref{eq:loss_gan}, resulting in the loss function $\mathcal{L}^{D} = \mathcal{L}^{D}_{GAN}(G,D)$. 

\section{Experiments} \label{sec:quant_experiments}

In this section, we report quantitative and qualitative experimental results. Dataset and evaluation details are provided, with an evaluation of our adversarial \ac{slp} model to follow.

\subsection{Implementation Details}
\paragraph{Dataset:} \label{sec:dataset}
We evaluate our approach on the publicly available \ac{ph14t} dataset introduced by Camgoz et al. \cite{camgoz2018neural}. The corpus provides 8257 German sentences and sign gloss translations alongside parallel sign pose videos of a combined 835,356 frames. We train our adversarial model to generate sign pose sequences of skeleton joint positions. Manual features of each video are extracted in 2D using OpenPose \cite{cao2018openpose}, and lifted to 3D using the skeletal model estimation improvements presented in \cite{zelinka2020neural}. For non-manual features, we represent facial landmarks as 2D coordinates, again extracted using OpenPose \cite{cao2018openpose}. The face coordinates are scaled to a consistent size and then centered around the nose joint. Each frame is then represented by the normalised joints of the signer, as $x$, $y$ and $z$ coordinates. 

\paragraph{Implementation setup:}

We setup our adversarial training with a progressive transformer generator built with 2 layers, 4 heads and a 512 embedding size. Our discriminator consists of 3 1D convolution layers, each with a feature size of 64 and a filter size of 10. We jointly train $G$ and $D$ by providing batches of source spoken language and target sign pose sequences, updating the model weights simultaneously with their respective loss functions $\mathcal{L}^{G}$ and $\mathcal{L}^{D}$. Experimentally, we find the best generator loss weights to be $\lambda_{Reg}=100$ and $ \lambda_{GAN}=0.001$.

During testing, we drop $D$ and use the trained $G$ to produce sign pose sequences given an input text. All parts of our network are trained with Xavier initialisation \cite{glorot2010understanding} and Adam optimization \cite{kingma2014adam}, with a learning rate of $10^{-3}$. Our code is based on Kreutzer et al.'s NMT toolkit, JoeyNMT \cite{JoeyNMT}, and implemented using PyTorch \cite{paszke2017automatic}.

\paragraph{Evaluation:}

We use the back translation evaluation metric for \ac{slp} introduced by Saunders~\etal \cite{saunders2020progressive}, employing a pre-trained \ac{slt} model \cite{camgoz2020sign} to translate the produced sign pose sequences back to spoken language. This is likened to the use of inception score for generative models \cite{salimans2016improved}, using a pre-trained classifier. BLEU and ROUGE scores are computed against the original input, with BLEU n-grams from 1 to 4 provided for completeness. The \ac{slp} evaluation protocols on the \ac{ph14t} dataset, set by \cite{saunders2020progressive}, are as follows: \textbf{Gloss to Pose (G2P)} is the production of sign pose from gloss intermediary, evaluating the sign production capabilities; \textbf{Text to Pose (T2P)} is the production of sign pose directly from spoken language, requiring both a translation to sign representation and a subsequent production of sign pose. 

\subsection{Adversarial Training}

We start with evaluation of our proposed adversarial training regime, initially producing only manual features to isolate the effect of the adversarial loss. We first conduct experiments on the \textbf{Gloss2Pose (G2P)} task, evaluating the production capabilities of our network. As shown in Table~\ref{tab:gloss_to_pose}, our adversarial training regime improves performance over Saunders \etal, a model trained solely with a regression loss \cite{saunders2020progressive}. This shows that the inclusion of a discriminator model increases the comprehension of sign production. We believe this is due to the discriminator pushing the generator towards a more expressive and articulate production, in order to deceive the adversary. This, in turn, increases the sign content contained in the generated sequence, leading to a more understandable output.

We next experiment with conditioning the discriminator on the source input, to provide discrimination upon both the raw translation and the realism of sign production. As shown, the additional conditioning on the source input improves performance even further. We believe this is due to the generator now requiring a more accurate translation to fool the discriminator, improving the mapping between source input and sign pose.

\begin{table}[t!]
\centering
\resizebox{0.9\linewidth}{!}{%
\begin{tabular}{@{}p{2.8cm}ccccc|ccccc@{}}
\toprule
 & \multicolumn{5}{c}{DEV SET} & \multicolumn{5}{c}{TEST SET} \\ 
\multicolumn{1}{c|}{Configuration:} & BLEU-4         & BLEU-3         & BLEU-2         & BLEU-1         & ROUGE          & BLEU-4     & BLEU-3         & BLEU-2         & BLEU-1         & ROUGE          \\ \midrule

\multicolumn{1}{r|}{Regression \cite{saunders2020progressive}}     &  11.93 & 15.08 & 20.50 & 32.40 & 34.01 & 10.43  & 13.51 &  19.19 & 31.80 & 32.02 \\ 

\multicolumn{1}{r|}{\textbf{Adversarial (Ours)}} & 12.63  & 15.83 & 21.37 & 32.94 & 35.11 & 11.63 & 14.78 & 20.49 & 32.70 & 33.47  \\
\multicolumn{1}{r|}{\textbf{Conditional Adv. (Ours)}} & {\B 12.74}  & {\B 15.97} & {\B 21.68} & {\B 33.95} & {\B 35.83} & {\B 11.70} & {\B 14.95} & {\B 20.86} & {\B 33.51} & {\B 33.64} \\ \bottomrule
\end{tabular}%
}
\caption{Adversarial Training results on the Gloss to Pose (G2P) task}
\label{tab:gloss_to_pose}
\end{table}

Our next experiment evaluates the performance of our adversarial training approach for the \textbf{Text2Pose (T2P)} task. Table~\ref{tab:text_to_pose} demonstrates that our adversarial model again achieves state-of-the-art results, further showcasing the effect of adversarial training. As the discriminator is conditioned upon the source text, the generator is prompted to accomplish both the accurate translation and realistic production tasks simultaneously.

\begin{table}[t!]
\centering
\resizebox{0.9\linewidth}{!}{%
\begin{tabular}{@{}p{3.6cm}ccccc|ccccc@{}}
\toprule
 & \multicolumn{5}{c}{DEV SET} & \multicolumn{5}{c}{TEST SET} \\ 
\multicolumn{1}{c|}{Configuration:} & BLEU-4         & BLEU-3         & BLEU-2         & BLEU-1         & ROUGE          & BLEU-4         & BLEU-3         & BLEU-2         & BLEU-1         & ROUGE          \\ \midrule

\multicolumn{1}{r|}{Regression \cite{saunders2020progressive}}     &  11.82 & 14.80 & 19.97 & 31.41 & 33.18 & 10.51 & 13.54 & {\B 19.04} & {\B 31.36} & 32.46 \\
\multicolumn{1}{r|}{\textbf{Conditional Adv. (Ours)}} & {\B 12.65} & {\B 15.61} & {\B 20.58} & {\B 31.84} & {\B 33.68} & {\B 10.81} & {\B 13.72} & 18.99 & 30.93 & {\B 32.74} \\ \bottomrule
\end{tabular}%
}
\caption{Adversarial Training results on the Text to Pose (T2P) task}
\label{tab:text_to_pose}
\end{table}

\subsection{Multi-Channel Sign Production}
Our final experiment evaluates the production of non-manual features, either independently (Non-M), or in combination with manual features (M + Non-M). We first produce sign using a sole regression loss and subsequently add the proposed adversarial loss, with G2P results shown in Table~\ref{tab:non_manual}. The sole production of non-manual features contains less signing information than manuals, shown by the relatively low BLEU-4 score of 7.39. This is because facial features complement the manual communication of the hands, providing contextual syntax to emphasise meaning as opposed to independently delivering content.

However, the combination of manual and non-manual feature production significantly increases performance to the highest BLEU-4 score of 13.16. Even the regression model improves performance compared to the manual production alone, highlighting the isolated effect. We believe the multi-channel sign production allows the communication of complementary information, with non-manuals providing further context to increase comprehension. This results in an articulate sign production, moving the field of \ac{slp} closer towards a more understandable output. The addition of adversarial training further improves the performance of both non-manual and manual feature production, indicating the ability of our approach to capture the full content of the sign and its morphology.

\begin{table}[t!]
\centering
\resizebox{0.9\linewidth}{!}{%
\begin{tabular}{@{}p{2.8cm}ccccc|ccccc@{}}
\toprule
 & \multicolumn{5}{c}{DEV SET} & \multicolumn{5}{c}{TEST SET} \\ 
\multicolumn{1}{c|}{Configuration:} & BLEU-4  & BLEU-3 & BLEU-2 & BLEU-1 & ROUGE & BLEU-4 & BLEU-3 & BLEU-2 & BLEU-1 & ROUGE \\ \midrule
\multicolumn{1}{r|}{Regression (Non-M)} & 7.19  & 9.13  & 12.93 & 23.31 & 25.01 & 6.51 & 8.50 & 12.44 & 23.85 & 24.38 \\ 
\multicolumn{1}{r|}{Adversarial (Non-M)} & 7.39  & 9.38 & 13.35 & 24.38 & 25.65 & 7.12 & 9.10 & 13.02 & 24.40 & 25.16 \\ \midrule
\multicolumn{1}{r|}{Regression (M + Non-M)}  & 12.12 & 15.38 & 20.97 & 32.67 & 35.21 & 11.54 & 14.53 & 20.05 & 31.63 & {\B 34.22} \\
\multicolumn{1}{r|}{Adversarial (M + Non-M)} & {\B 13.16} & {\B 16.52} & {\B 22.42} & {\B 34.09} & {\B 36.75} & {\B 12.16} & {\B 15.31} & {\B 20.95} & {\B 32.41} & 34.19 \\ \bottomrule
\end{tabular}%
}
\caption{Non-Manual production results on the G2P task (Non-M: Non-Manual, M: Manual)}
\label{tab:non_manual}
\end{table}
\begin{figure}[b!]
\centering
    \includegraphics[width=0.95\linewidth]{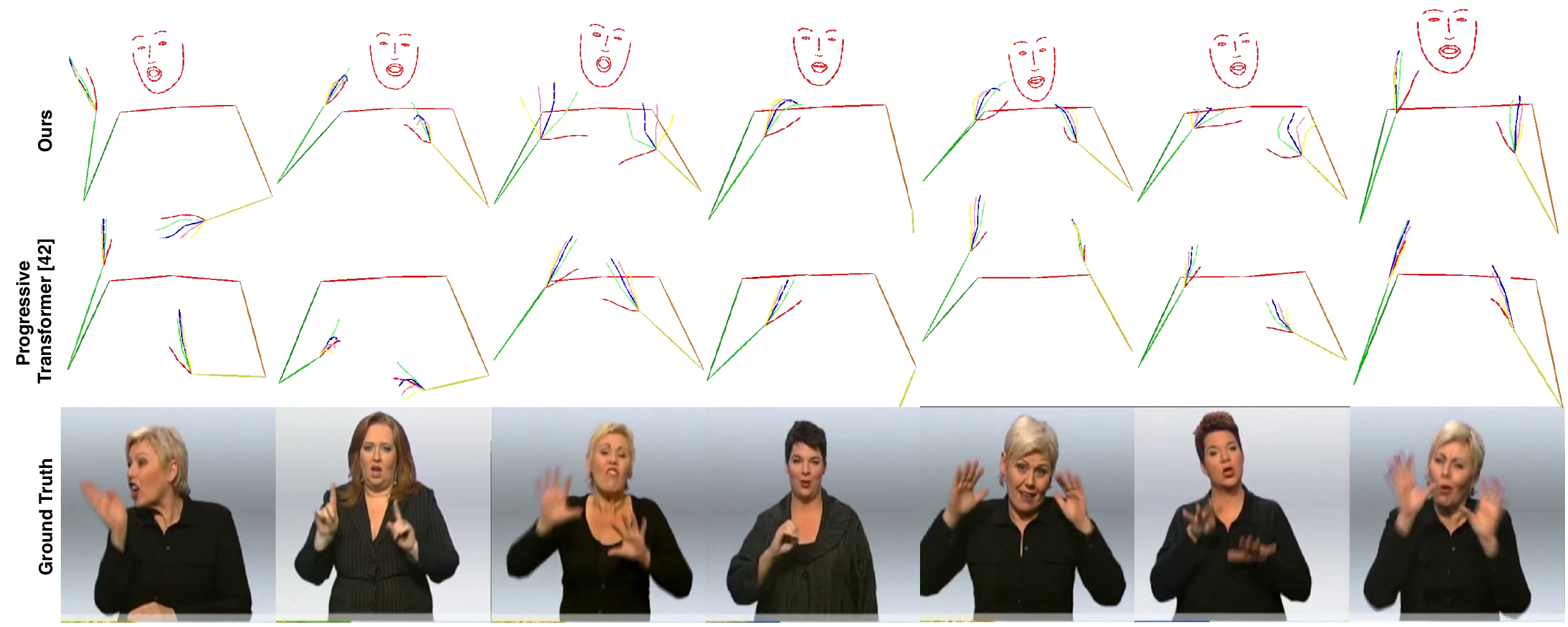}
    \caption{Produced sign pose examples from our proposed model (top) compared to that of \cite{saunders2020progressive} (middle), alongside the ground truth frame (bottom)}
    \label{fig:qualitative_output_Prog}
\end{figure}%
\begin{figure}[]
\centering
    \includegraphics[width=0.95\linewidth]{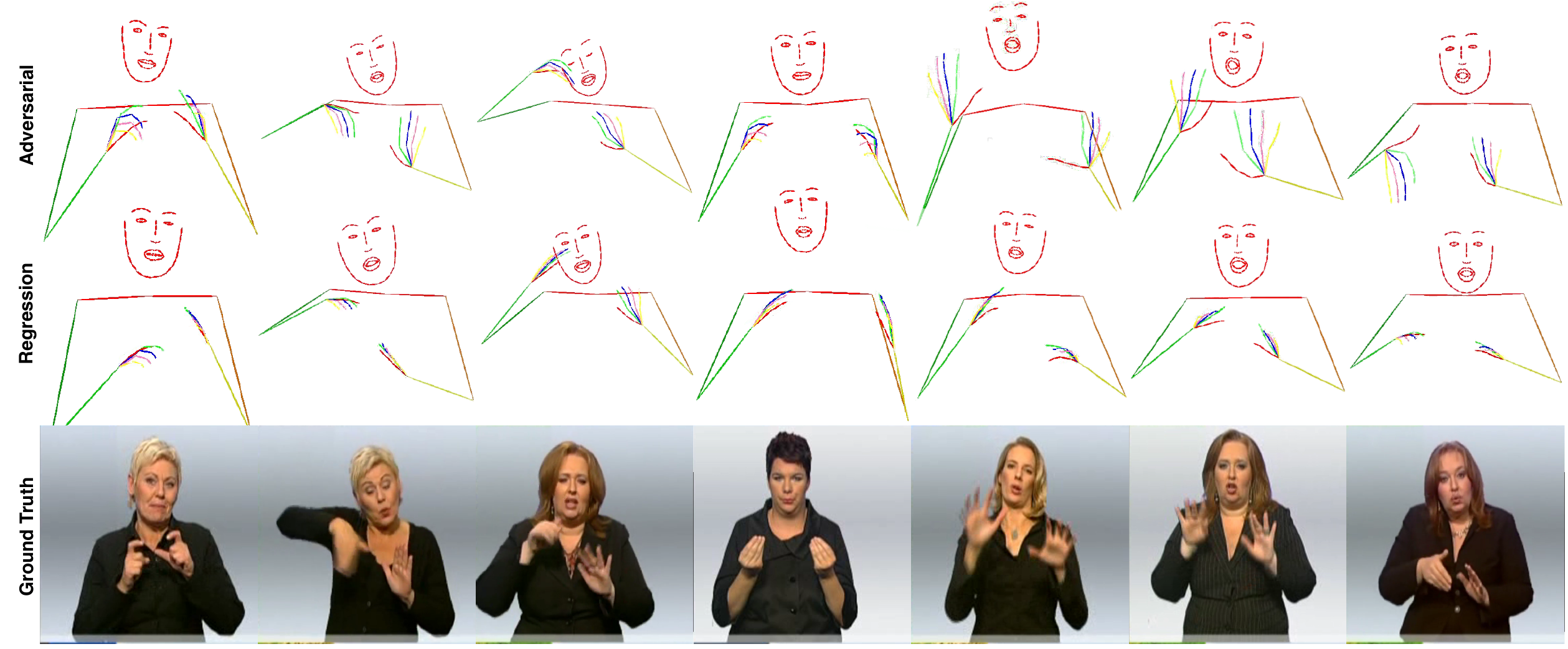}
    \caption{Produced sign pose examples from our proposed adversarial model (top) and a regression model comparison (middle), alongside the ground truth frame (bottom)}
    \label{fig:qualitative_output}
\end{figure}%

\subsection{Qualitative Experiments} \label{sec:qual_experiments}

Figure~\ref{fig:qualitative_output_Prog} shows example frames of multi-channel sign pose sequences produced by our proposed adversarial training approach, compared against Saunders \etal \cite{saunders2020progressive}. The examples show an increase in articulation and realism, with a highlight on the importance of non-manual production. Specific to non-manual features, we find a close correspondence to the ground truth video alongside accurate mouthings and head movements.

Figure~\ref{fig:qualitative_output} shows the isolated effect of adversarial training compared to a pure regression approach. Viewed alongside ground truth frames, the produced sign pose demonstrates accurate manual and non-manual production. We find that the addition of adversarial training produces sequences of increased articulation, with a smoother production. Hand shapes can be seen to be more expressive and meaningful, an important result for sign comprehension and understandable \ac{slp}. Further examples are available in the supplementary materials.

%
\section{Conclusion} \label{sec:conclusion}
Sign languages are visual multi-channel languages and the principal form of communication of the Deaf. \acf{slp} requires the production of the full sign morphology in an articulate manner in order to be understood by the Deaf community. Previous deep learning based \ac{slp} work has generated only manual features, in an under-expressed production due to the problem of regression to the mean.

In this paper, we proposed an adversarial multi-channel approach for \ac{slp}. Framing \ac{slp} as a minimax game, we presented a conditional adversarial discriminator that measures the realism of generated sign sequences and pushes the generator towards an articulate production. We also introduced non-manual feature production to fully encapsulate the sign language articulators. We evaluated on the \ac{ph14t} dataset, showcasing the effectiveness of our adversarial approach by reporting state-of-the-art results for manual production and setting baselines for non-manuals. 

As future work, we would like to further increase the realism of sign production by generating photo-realistic human signers, using \ac{gan} image-to-image translation models \cite{isola2017image,zhu2017unpaired,chan2019everybody} to expand from the skeleton representation. Furthermore, user studies in collaboration with the Deaf are required to evaluate the reception of the produced sign pose sequences.

\section{Acknowledgements}
This work received funding from the SNSF Sinergia project `SMILE' (CRSII2 160811), the European Union's Horizon2020 research and innovation programme under grant agreement no. 762021 `Content4All' and the EPSRC project `ExTOL' (EP/R03298X/1). This work reflects only the authors view and the Commission is not responsible for any use that may be made of the information it contains. We would also like to thank NVIDIA Corporation for their GPU grant.

\bibliography{bibliography}
\end{document}